\pdfoutput=1

\documentclass[11pt]{article}

\usepackage{ACL2023}

\usepackage{times}
\usepackage{latexsym}

\usepackage[T1]{fontenc}

\usepackage[utf8]{inputenc}

\usepackage{microtype}

\usepackage{inconsolata}

\usepackage{array}
\usepackage{multirow}
\usepackage{makecell}
\usepackage{soul}
\usepackage{url}
\usepackage{graphicx}
\usepackage{amsmath}
\usepackage{booktabs}
\usepackage{subfigure}
\usepackage{algorithmic}
\usepackage{enumitem}
\usepackage{amsthm,amsmath,amssymb}
\usepackage{mathrsfs}
\usepackage{bm}
\usepackage{hyperref}
\usepackage[ruled,linesnumbered]{algorithm2e}

\title{Pre-training Multi-party Dialogue Models with Latent Discourse Inference}

\author{Yiyang Li$^{1,2,}$\thanks{\; Work done while interning at Tencent AI Lab.} , Xinting Huang$^{3,\dag}$, Wei Bi$^{3}$ \and Hai Zhao$^{1,2,}$\thanks{\; Corresponding author. This paper was partially supported by Key Projects of National Natural Science Foundation of China (U1836222 and 61733011).} \\
        $^1$ Department of Computer Science and Engineering, Shanghai Jiao Tong University\\
        $^2$ Key Laboratory of Shanghai Education Commission for Intelligent Interaction\\and Cognitive Engineering, Shanghai Jiao Tong University\\
        $^3$ NLP Center, Tencent AI-Lab\\
        \texttt{eric-lee@sjtu.edu.cn, zhaohai@cs.sjtu.edu.cn}\\
        \texttt{\{timxinhuang,victoriabi\}@tencent.com}
}

\begin{document}
\maketitle
\begin{abstract}
Multi-party dialogues are more difficult for models to understand than one-to-one two-party dialogues, since they involve multiple interlocutors, resulting in interweaving reply-to relations and information flows. To step over these obstacles, an effective way is to pre-train a model that understands the discourse structure of multi-party dialogues, namely, to whom each utterance is replying. However, due to the lack of explicitly annotated discourse labels in multi-party dialogue corpora, previous works fail to scale up the pre-training process by putting aside the unlabeled multi-party conversational data for nothing. To fully utilize the unlabeled data, we propose to treat the discourse structures as latent variables, then jointly infer them and pre-train the discourse-aware model by unsupervised latent variable inference methods. Experiments on multiple downstream tasks show that our pre-trained model outperforms strong baselines by large margins and achieves state-of-the-art (SOTA) results, justifying the effectiveness of our method.
The official implementation of this paper is available at \url{https://github.com/EricLee8/MPD_EMVI}.
\end{abstract}

\section{Introduction}
\label{sec:intro}
Dialogue system is an important area that has been studied for a long time in natural language processing field. Different from plain texts, dialogues are harder for models to understand since they are full of informal, colloquial expressions, and many ellipses \cite{yang-choi-2019-friendsqa, reddy-etal-2019-coqa, li-etal-2022-back}. Among them, multi-party dialogues are even more complex since they involve multiple interlocutors, resulting in interweaving reply-to relations and information flows \cite{MPC-BERT, discourse-EMRC, who-says-what-to-whom}. Specifically, in multi-party dialogues, the current utterance can be a reply to any preceding utterance in the dialogue history, forming complex discourse structures.

Intuitively, it is important for models to perceive the discourse structures, or in other words, to whom each utterance is replying, when comprehending multi-party dialogues. This intuition is in line with the process we humans participate in multi-party dialogues: we first read or listen to the dialogue history, knowing who speaks what to whom, then choose an utterance as the addressee, and finally utter a response. Literature has also justified that incorporating the discourse knowledge into models is beneficial for better understanding multi-party dialogues \cite{li-etal-2020-molweni, thread-kenny, self-and-pseudo, maxb-disentangle}. Unfortunately, the process of choosing addressees is a naturally unobservable action, resulting in a large amount of multi-party conversational data without addressee labels. In this work, we focus on leveraging the unlabeled data to pre-train a model for multi-party dialogue understanding.

To utilize the discourse structure, previous works seek help from human laborers to annotate the addressee labels on small datasets, where they either explicitly model the discourse structure using Graph Neural Networks or multi-task learning \cite{GSN,discourse-EMRC,DADGraph,he-etal-2021-multi,gu-etal-2022-hetermpc}, or attempt to pre-train a model using objectives that are related to addressees by supervised learning \cite{MPC-BERT}. These works heavily rely on annotated addressee labels, which are rare in practice since the annotation process requires large amounts of human resources. As a result, they fail to be practical in real-world applications and are hard to scale up by utilizing more unlabeled multi-party conversational data.

To make full use of the unlabeled corpora, a natural idea is to treat the unobservable discourse structure (reply-to relations) as latent variables, then adopt latent variable models to jointly infer them and optimize the discourse-aware models. However, it is not that simple when it comes to practice. For the Expectation-Maximization (EM) algorithm, the posterior distribution of the reply-to relations is intractable since it requires a square-level time complexity. If we turn to Variational Inference (VI) for help, the choice of the categorical prior distribution of the reply-to relations becomes troublesome: naive assumptions such as uniform distributions are too weak to make the training process converge.

To step over the above obstacles, we subtly combine the single-turn EM algorithm and multi-turn VI into a two-stage pre-training strategy. In the first stage, we adopt the EM algorithm to jointly model the context-response matching objective and single-turn addressee inference, which requires only a linear time complexity and can preliminarily guide the model to a relatively good converging point with utterance-level knowledge. In the second stage, we extend the latent variables from single-turn addressees to multi-turn reply-to relations and optimize the model via both the EM algorithm and VI framework, where the prior distribution of the reply-to relations is no longer troublesome since it can be derived exactly from the E-steps. This stage further enhances the model with discourse-level knowledge and guides it converge to a better point.

To sum up, the contributions of this work are:
\begin{itemize}[leftmargin=*, topsep=1pt]
    \setlength{\itemsep}{0pt}
    \setlength{\parsep}{0pt}
    \setlength{\parskip}{0pt}
    \item We successfully scale up the pre-training for multi-party dialogue understanding by leveraging the huge amounts of multi-party conversational corpora without addressee labels, while previous methods fail to work on these corpora.
    \item We subtly combine the single-turn EM algorithm and multi-turn VI framework in a two-stage pre-training process, which equips the model with knowledge of different granularities and makes it converge to an ideal point.
    \item The pre-trained model serves as a powerful encoder for multi-party dialogues and outperforms strong baselines by large margins, achieving SOTA results on multiple downstream tasks.
\end{itemize}

\section{Related Works}
\subsection{Multi-party Dialogue Modeling}
\label{sec:multi-party-modeling}
Several works have studied the modeling of multi-party dialogues before. \citet{GSN} propose to encode the reply-to relations with Graph Structural Networks (GSN). They utilize the addressee annotations and speaker information in the dataset to construct discourse and speaker graphs, then adopt a backward-forward strategy to pass messages between utterances. \citet{discourse-EMRC,gu-etal-2022-hetermpc} further extend the modeling from homogeneous graphs to heterogeneous graphs by utilizing the Relational Graph Convolutional Networks to encode the heterogeneous information. However, their solutions all require annotated addressee labels in the multi-party dialogue dataset, which are rare and expensive to obtain in real-world applications. On the contrary, our work requires no addressee annotations, which saves human labors and can be scaled up using large unlabeled corpora.

Most related to our work, \citet{li2023em} attempts to improve the response generation model for multi-party dialogues by employing the EM algorithm to infer single-turn addressees. However, their approach encounters limitations when it comes to expanding the pre-training process due to the slow generative E-steps. Additionally, their work fails to fully exploit the discourse structure of the dialogue history, as they solely focus on the single-turn addressees. In contrast, our method not only scales up the pre-training by employing faster objectives, but also extends the latent variables from single-turn addressees to multi-turn reply-to relations to enhance the model with discourse-level knowledge, which is more important in comprehending multi-party conversations.

\subsection{Dialogue Pre-training}
\label{sec:dialog_pretrain}
To bridge the gap between pre-trained language models (PLMs) on plain texts and dialogue texts, many attempts have been made to pre-train a model for dialogues. \citet{bao-etal-2020-plato, chen-etal-2022-dialogved} treat the dialogue intent as discrete or continuous latent variables to pre-train a model that solves the one-to-many problem in dialogue response generation task. \citet{mehri-etal-2019-pretraining,xu-zhao-2021-dialogue,zhang-zhao-2021-structural} design different self-supervised objectives for two-party dialogue context modeling. Different from their two-party setting, our work focuses on the multi-party scenario, where the addressee information should be concerned. \citet{MPC-BERT} also consider pre-training a model for multi-party dialogue understanding. They pre-train their model on a small dataset with annotated addressee labels by supervised addressee-related objectives. Since annotations are required, their pre-training strategy fails to scale up by using the unlabeled data. In contrast, our method is labor-free since the addressees are inferred by unsupervised latent-variable methods.

\begin{figure*}[tbp]
    \centering
    \includegraphics[width=0.83\textwidth]{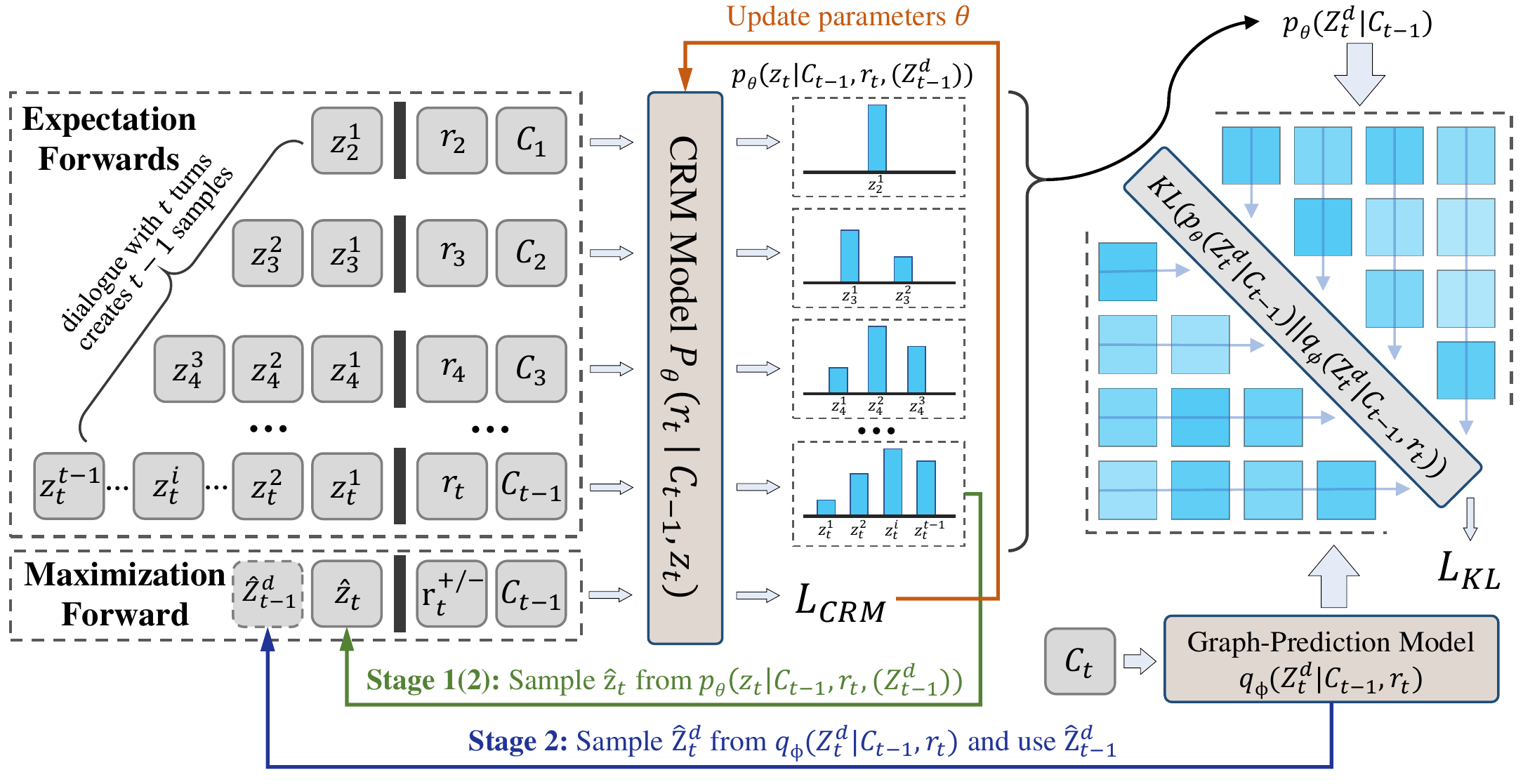}
    \caption{The overview of our pre-training process. The left part shows the turn-level Expectation-Maximization process while the right part illustrates the discourse-level Variational Inference enhancement.}
    \label{fig:model_overview}
\end{figure*}

\section{Methodology}
In general, Figure \ref{fig:model_overview} illustrates the overview of the proposed two-stage pre-training strategy.
The left part illustrates the single-turn Expectation-Maximization process, where we iteratively conduct E-steps to infer the latent addressee $z_t$ (left-upper part and the green arrow), and M-steps to optimize the model via addressee-aware context-response matching (CRM) objective (left-lower part and the orange arrow).
The right part illustrates the multi-turn Variational Inference process, which is incorporated into the EM framework in the second pre-training stage. We extend the latent variables from the single-turn addressees to multi-turn addressee-graphs, and jointly optimize the discourse-aware context-response matching model (the blue arrow) and the graph-prediction model $q_{\bm{\phi}}$ by Variational Inference.
In the next sections, we will introduce the two pre-training stages in detail.

\subsection{Single-turn Addressee Inference}
\label{sec:EM}
As mentioned in Section \ref{sec:intro}, simply applying the EM algorithm to infer all reply-to relations in the dialogue requires a square-level time complexity, which is intolerably time-consuming for the pre-training on large corpora. To solve this issue, we step back in the first pre-training stage to focus on the modeling and inference of single-turn addressees. For one thing, it requires only a linear time complexity for each training instance and hence can be optimized via the EM algorithm. For another, the addressee distributions output by the E-steps can derive the prior distribution of the reply-to relations, which can be utilized by the Variational Inference process in the second pre-training stage.

\subsubsection{Preliminaries}
\label{sec:EM_overview}
Let's consider the process that humans participate in a multi-party dialogue in the $t_{th}$ turn: we first read the dialogue history $C_{t-1}$, then choose an addressee utterance $z_t$ that we want to reply, and finally utter a response sentence $r_t$. Formally, a multi-party dialogue corpus contains dialogues with format $(C_{t-1}, z_t, r_t)$, where the annotations of $z_t$ are lacking in most corpora. Here $C_{t-1} = \{\operatorname{S_1: U_1 [SEP] S_2: U_2 [SEP]\dots S_{t-1}: U_{t-1} [SEP] S_t}\}$, where $\operatorname{S_i}$ and $\operatorname{U_i}$ are the speaker and utterance of the $i_{th}$ turn, respectively. Addressee $z_t\in [1, t-1]$ is a one-hot vector that indicates to whom we reply in the current turn $t$. In our settings, each utterance except the first one has exactly one addressee.

The conversation process can be formulated as $p_{\bm{\theta}}(r_t | z_t, C_{t-1})$, which models the probability of $r_t$ being the correct response given $C_{t-1}$ and $z_t$ under trainable parameters $\bm{\theta}$. In large datasets without addressee labels $z_t$, we should infer the unobservable latent addressees. To this end, we adopt the EM algorithm to iteratively infer the addressees $p_{\bm{\theta}}(z_t | C_{t-1}, r_t)$ during the E-steps, and optimize the model $p_{\bm{\theta}}(r_t | z_t, C_{t-1})$ using the CRM objective during the M-steps.

\subsubsection{Maximization Step}
\label{sec:mstep}
Suppose we have already obtained the inferred addressees from the E-step, two questions should be answered in the M-step: how to design the addressee-aware model architecture, and how to design the CRM task that enforces the model to leverage addressee information.

To answer the first question, our solution is straightforward but effective: similar to the speaker or turn embeddings in previous works \cite{speaker-bert, dialog-bert}, we add an addressee embedding on top of the token and positional embeddings to indicate which utterance is the current addressee. Note that we have also tried other addressee modeling methods such as the prompt-based ones, yet they are not as effective as the addressee embeddings.

To answer the second question, we first follow the common practice to formulate the CRM task as a binary classification problem \cite{CRM-ijcai, CRM-su}, where the model should distinguish positive (correct) responses $r_t^+$ from the negative ones $r_t^-$ in the current dialogue turn $t$. To make the CRM task more addressee-related, besides simple negatives that are randomly sampled from the whole training corpus, we also construct hard negatives that are sampled from the later ($>t$ turns) utterances in the same dialogue. \citet{liu2019roberta} point that simple negatives are easily distinguishable from positive ones by their topic differences. In other words, they can be predicted as negatives without the specified addressee information, which can not help the addressee inference process in the E-step. In contrast, the topic of each hard negative response is coherent with the current dialogue, making them hard to be classified with only the topic or sequential features. As a result, the model is forced to seek clues from the speaker and addressee information to distinguish those hard negatives, which greatly benefits the E-step.

With the model and training data at hand, we adopt binary cross-entropy loss as the objective function for the CRM task:
\begin{equation}
    \label{eq:crm_loss}
    \begin{aligned}
        \mathcal{L}_{CRM} &= -(y_t\times \operatorname{log[}p_{\bm{\theta}}(r_t | z_t, C_{t-1})\operatorname{]}\\
        &+(1-y_t)\times \operatorname{log[}1-p_{\bm{\theta}}(r_t | z_t, C_{t-1})\operatorname{]})
    \end{aligned}
\end{equation}
Here $y_t\in \{0, 1\}$ is the ground truth label that indicates whether $r_t$ is a positive response. The left lower part and the orange arrow of Figure \ref{fig:model_overview} illustrate the maximization step, where we ignore $\hat{Z}_{t-1}^d$ since it will be introduced in Section \ref{sec:vi}.

\subsubsection{Expectation Step}
\label{sec:estep}
The inference of latent addressees can be formulated as calculating $p_{\bm{\theta}}(z_t | C_{t-1}, r_t)$. In other words, given the dialogue history $C_{t-1}$ and current response $r_t$, we should infer the posterior categorical distribution of the addressee $z_t\in [1, t-1]$. Consider the factorization of this posterior distribution:
\begin{equation}
    \begin{aligned}
        &p_{\bm{\theta}}(z_t | C_{t-1}, r_t) = \frac{p_{\bm{\theta}}(C_{t-1}, z_t, r_t)}{p_{\bm{\theta}}(C_{t-1}, r_t)}\\
        &= \frac{p_{\bm{\theta}}(C_{t-1})\times p_{\bm{\theta}}(z_t|C_{t-1})\times p_{\bm{\theta}}(r_t|z_t, C_{t-1})}{p_{\bm{\theta}}(C_{t-1})\times p_{\bm{\theta}}(r_t|C_{t-1})}\\
        &= \frac{p_{\bm{\theta}}(z_t|C_{t-1})\times p_{\bm{\theta}}(r_t|z_t, C_{t-1})}{p_{\bm{\theta}}(r_t|C_{t-1})}
    \end{aligned}
\end{equation}
where the factorization order of the numerator follows human habits when participating in a multi-party dialogue mentioned at the beginning of Section \ref{sec:EM_overview}. In the denominator, $p_{\bm{\theta}}(r_t|C_{t-1})$ is irrelevant to $z_t$. In the numerator, we assume a uniform prior distribution $p_{\bm{\theta}}(z_t|C_{t-1})$, hence this term is also irrelevant to $z_t$. Hence, we can derive that:
\begin{equation}
    \label{eq:propto}
    p_{\bm{\theta}}(z_t | r_t, C_{t-1}) \propto p_{\bm{\theta}}(r_t|z_t, C_{t-1})
\end{equation}
Adopting this equation and the trained CRM model $p_{\bm{\theta}}(r_t|z_t, C_{t-1})$ from the M-step, we can now calculate the posterior distribution of $z_t$ by traversing all possible addressees $\{z_t^i\}_{i=1}^{t-1}$:
\begin{equation}
    \label{eq:addr_inference}
    p_{\bm{\theta}}(z^i_t|r_t, C_{t-1}) = \frac{p_{\bm{\theta}}(r_t|z^i_t, C_{t-1})}{\sum_{j=1}^{t-1} p_{\bm{\theta}}(r_t|z^j_t, C_{t-1})}
\end{equation}
The left upper part and green arrow in Figure \ref{fig:model_overview} shows the E-step, where we ignore $Z^d_{t-1}$ since it will be introduced in Section \ref{sec:vi}.

\subsection{Multi-turn Addressee-graph Inference}
\label{sec:vi}
Once the EM iterations have reached a relatively good converging point, we dive into the second stage of training by additionally integrating the multi-turn Variational Inference task into the EM framework. This stage further enhances the model with discourse-level knowledge, making it possible to converge to a better point.

The discourse-level VI extends the latent variables from single-turn addressees $z_t$ to multi-turn addressee-graphs $Z_t^d\in \mathcal{R}^{t\times t}$, which is an adjacent matrix indicating to which addressee each utterance is replying to. In other words, the model now should infer all the addressees of each utterance $U_i$ in the dialogue context $C_t$. As mentioned in Section \ref{sec:EM}, adopting the EM algorithm to infer $Z_t^d$ is intolerably time-consuming. To solve this issue, we borrow the idea of Variational Inference \cite{VAE} to adopt a graph-prediction model $q_{\bm{\phi}}(Z_t^d|C_{t-1}, r_t)$ with additional trainable parameters $\bm{\phi}$ to predict the addressee-graphs. Formally, we maximize the log-likelihood of the observed data $\operatorname{log}p_{\bm{\theta}}(r_t|C_{t-1})$ (conditioned on the dialogue history $C_{t-1}$) by improving its Evidence Lower Bound (ELBO):
\begin{equation}
    \begin{aligned}
        &\operatorname{ELBO}(\bm{\theta}, \bm{\phi}; r_t, C_{t-1}) = \\
        & \quad \mathbb{E}_{q_{\bm{\phi}}(Z_t^d|r_t, C_{t-1})}[\operatorname{log}p_{\bm{\theta}}(r_t|Z_t^d, C_{t-1})]\\
        & \quad -D_{KL}(q_{\bm{\phi}}(Z_t^d|r_t, C_{t-1}) \Vert p_{\bm{\theta}}(Z_t^d|C_{t-1}))
    \end{aligned}
\end{equation}
Three important distributions are presented in this equation. First, $p_{\bm{\theta}}(r_t|Z_t^d, C_{t-1})$ is a new formulation of the CRM task, where single-turn addressees $z_t$ now becomes multi-turn addressee-graphs $Z_t^d$. Second, $p_{\bm{\theta}}(Z_t^d|C_{t-1})$ is the conditional prior distribution of latent variable $Z_t^d$ under parameters $\bm{\theta}$. Finally, $q_{\bm{\phi}}(Z_t^d|C_{t-1}, r_t)$ is the graph-prediction model, which predicts the edges from each response to its addressee by outputting the estimated posterior distribution of $Z_t^d$. Next, we introduce the modeling of these distributions in detail.

\subsubsection{Discourse-aware CRM}
Let's start with $p_{\bm{\theta}}(r_t|Z_t^d, C_{t-1})$. Given the dialogue history $C_{t-1}$ and the addressee-graph $Z_t^d$ sampled from $q_{\phi}$, we model the CRM task by imitating \emph{careful} human readers: when we \emph{seriously} reply to an utterance in a multi-party dialogue, instead of focusing solely on the current addressee utterance $z_t$ itself, we tend to focus more on the utterances in the reply-chain of $r_t$, namely, the $k$-hop ancestors of $r_t$ in the addressee-graph $Z_t^d$. Formally, we first extract the utterance representations of the $k$-hop ancestors of $r_t$ to form a reply-chain information representation $H_t^k\in \mathcal{R}^{k\times d}$, then model $p_{\bm{\theta}}(r_t|Z_t^d, C_{t-1})$ with an MLP.

To accelerate the computation of the $k$-hop ancestors, we construct a one-hot vector $a_t\in \mathcal{R}^{1\times t}$ to indicate the position of the current response $r_t$. Right-multiplying this vector by the addressee-graph matrix $Z_t^d$ for $i$ times yields the position vector of its $i_{th}$ ancestor. $p_{\bm{\theta}}(r_t|Z_t^d, C_{t-1})$ can now be formulated as follows:
\begin{equation}
    \begin{aligned}
        & H_t^k = \operatorname{concat}[\{a_t (Z_t^d)^i\}_{i=0}^{k-1}]\cdot H_t^u \in \mathcal{R}^{k\times d}\\
        & p_{\bm{\theta}}(r_t|Z_t^d, C_{t-1}) = \sigma(\operatorname{MLP}_{\bm{\theta}}(\operatorname{flatten}(H_t^k)))
    \end{aligned}
\end{equation}
Here $\operatorname{concat}[\cdot]$ is concatenation, $\operatorname{flatten}$ means squeezing the matrix into a vector, $\operatorname{MLP}_{\bm{\theta}} \in \mathcal{R}^{kd\times 1}$ is a linear projection and $\sigma$ is the Sigmoid function. In this pre-training stage, $p_{\bm{\theta}}(z_t | r_t, C_{t-1})$ and $p_{\bm{\theta}}(r_t|z_t, C_{t-1})$ in the equations of Section \ref{sec:EM} have now become $p_{\bm{\theta}}(z_t | r_t, Z_{t-1}^d, C_{t-1})$ and $p_{\bm{\theta}}(r_t|Z_t^d, C_{t-1})$, respectively. For more detailed proofs, please refer to Appendix \ref{app:estep_stage2}.

\subsubsection{Conditional Prior Distribution}
Then, we focus on the conditional prior distribution $p_{\bm{\theta}}(Z_t^d|C_{t-1})$. The choice of the prior distribution is vital to the convergence of Variational Inference \cite{VAE, LoREN}. Previous works either make strong assumptions over the prior distribution, like Uniform and Gaussian \cite{qian-etal-2022-controllable}, or use additional annotation models to approximate the prior distribution \cite{LoREN}. However, as mentioned in Section \ref{sec:intro}, they fail to work in our scenario since naive assumptions are too weak to make the training process converge. Thanks to the EM training process, the prior distribution $p_{\bm{\theta}}(Z_t^d|C_{t-1})$ can be derived exactly from the previous $t-1$ E-steps in this dialogue. Formally, it can be calculated as:
\begin{equation}
    \label{eq:prior}
    \begin{aligned}
        & E(i) = p_{\bm{\theta}}(z_i|r_i, Z_{i-1}^d, C_{i-1})\\
        & p_{\bm{\theta}}(Z_t^d|C_{t-1}) = \Pi_{i=1}^{t-1} [E(i)]\cdot U(|z_t|)
    \end{aligned}
\end{equation}
Here $U(|z_t|)$ is a uniform distribution over the length of the candidates of $z_t$. Due to the page limit, we put the detailed derivations of this equation in Appendix \ref{app:prior}. This equation subtly combines the EM training framework and the VI process, which guides the model converge to a better point by incorporating accurate prior knowledge of the discourse-level addressee-graphs.

\subsubsection{Graph-prediction Model}
Finally, we end with the graph-prediction model $q_{\bm{\phi}}(Z_t^d|C_{t-1}, r_t)$. To compute the edges between each utterance pair, we first apply mean pooling over the corresponding token representations of each utterance to get utterance-level representations $H_t^u \in \mathcal{R}^{t\times d}$. After that, we compute the score of each utterance pair being the response-addressee by an MLP with trainable parameters $\bm{\phi}$ to get a scoring matrix $S^u\in \mathcal{R}^{t\times t}$. Finally, $q_{\bm{\phi}}$ is calculated as follows:
\begin{equation}
\label{eq:q-phi}
    q_{\bm{\phi}} = \operatorname{Gumbel-Softmax}(S^u+M^u)
\end{equation}
Here $M^u \in \mathcal{R}^{t\times t}$ is a masking matrix with $-\infty$ values on its upper triangular part to mask invalid positions, since each utterance can only reply to its previous ones. We adopt $\operatorname{Gumbel-Softmax}$ relaxation to make the sampling of $q_{\bm{\phi}}$ differentiable, following \citet{JangGP17, MaddisonMT17}.

\subsection{Pre-training Objectives}
\label{sec:total-loss}
Besides utterance-level CRM and discourse-level graph prediction, we also design an addressee-aware masked language modeling (MLM) task to preserve the token-level knowledge, which is introduced in detail in Appendix \ref{app:MLM}. To sum up, the overall training objective in the M-step is:
\begin{equation}
\label{eq:total-loss}
    \mathcal{L} = \mathcal{L}_{CRM} + \alpha \mathcal{L}_{KL} + \beta \mathcal{L}_{MLM}
\end{equation}
Here $\alpha$ and $\beta$ are two hyper-parameters and are set to $0$ at the first pre-training stage.

\section{Experiments}
In this section, we introduce the experimental settings and present the results on downstream tasks.

\subsection{Pre-training Settings}
\label{sec:pretrain-setting}
For the pre-training data, we use the script of \cite{zhang-etal-2020-dialogpt} to download Reddit posts from 2005 to 2020 and extract multi-party conversations to create a pre-training corpus of 17,154,613 dialogues. Since the pre-training corpus is huge, we split it into trunks of data and perform EM iterations on each of them. For backbone models, we choose BERT$_{\operatorname{base}}$ \cite{devlin2019bert} and ELECTRA$_{\operatorname{large}}$ \cite{clark2020electra}. The former takes $4$ days to converge in $8$ NVIDIA A100 GPUs and the latter takes $12$ days. For more details about the pre-training, please see Appendix \ref{app:process}.

\subsection{Downstream Settings}
To test the capability of our pre-trained model, we conduct experiments on four downstream tasks based on multi-party dialogues.

\textbf{Discourse Parsing} requires the model to parse the reply-to links (addressee-graphs) in a multi-party dialogue and classify their relation types at the same time. For this task, we adopt Molweni \cite{li-etal-2020-molweni} as the benchmark dataset and use the F1 score of graph-prediction (F1$_\text{G}$) and relation classification (F1$_\text{RL}$) as the evaluation metrics.

\textbf{Successful New Entry Prediction} is to predict whether a newcomer’s message will be responded to by other participants in a multi-party dialogue, which is formulated as a binary classification task. For this task, we adopt SNEP \cite{snep} as the benchmark dataset and use Area Under Curve (AUC) and F1 score as the evaluation metrics.

\textbf{Extractive Question Answering} requires the model to extract an answer span from the dialogue context given a question. For this task, we also adopt Molweni as the benchmark and use Exact-Match (EM) and F1 score as the evaluation metrics.

\textbf{Response Generation} aims at generating an appropriate response given the speaker and a specified addressee in a multi-party dialogue. For this task, we adopt Ubuntu IRC dataset \cite{GSN} as the benchmark dataset and use BLEU, METEOR, and ROUGE-L as the evaluation metrics.

For more details about the datasets (statistics, data sources, etc.), please refer to Appendix \ref{app:datasets}.

\begin{table*}
    \vspace{-0.5em}
    \centering
    \small
    \begin{tabular}{l|cc|cc|cc|cc}
        \specialrule{0.09em}{0.0pt}{0.2pt}
        \multirow{2}{*}{Model} & \multicolumn{2}{c|}{Discourse Parsing} & \multicolumn{2}{c|}{SNEP-Reddit} & \multicolumn{2}{c|}{SNEP-Twitter} & \multicolumn{2}{c}{Extractive Q.A.}\\
        \cline{2-9}
        & F1$_\text{RL}$ & F1$_\text{G}$ & AUC & F1 & AUC & F1 & EM & F1 \\
        \hline
        \emph{Adaptation Model} & & & & & & & & \\
        BERT-base & $61.06$ & $87.33$ & $63.89$ & $33.73$ & $81.50$ & $88.25$ & $47.78$ & $61.77$\\
        SPIDER-BERT & $62.79$ & $87.92$ & $64.88$ & $34.02$ & $81.98$ & $88.87$ & $48.69$ & $62.79$\\
        MPC-BERT & $63.91$ & $89.12$ & $65.08$ & $34.12$ & $82.56$ & $89.05$ & $47.29$ & $61.72$\\
        BERT+CRM & $63.08$ & $88.40$ & $67.06$ & $36.77$ & $83.61$ & $89.22$ & $49.66$ & $63.31$\\
        \quad +MLM & $63.79$ & $88.42$ & $67.32$ & $36.58$ & $83.72$ & $89.33$ & $50.03$ & $63.54$\\
        \quad \quad +VI & $\mathbf{64.97}$ & $\mathbf{90.31}$ & $\mathbf{68.16}$ & $\mathbf{36.97}$ & $84.06$ & $\mathbf{89.62}$ & $51.17$ & $\mathbf{64.89}$\\
        \hline
        \emph{Vanilla Model} & & & & & & & & \\
        BERT-base & $60.71$ & $87.45$ & $63.44$ & $32.57$ & $81.33$ & $87.85$ & $46.81$ & $60.20$\\
        SPIDER-BERT & $62.32$ & $87.68$ & $64.72$ & $33.32$ & $81.78$ & $88.75$ & $47.68$ & $61.16$\\
        MPC-BERT & $63.19$ & $88.75$ & $65.26$ & $34.63$ & $81.82$ & $88.83$ & $46.84$ & $60.11$\\
        BERT+CRM & $62.95$ & $88.17$ & $67.15$ & $35.88$ & $82.91$ & $89.11$ & $47.58$ & $61.74$\\
        \quad +MLM & $63.19$ & $88.91$ & $67.16$ & $36.36$ & $83.48$ & $88.92$ & $47.51$ & $62.43$\\
        \quad \quad +VI & $64.22$ & $89.59$ & $68.09$ & $36.96$ & $\mathbf{84.78}$ & $89.61$ & $\mathbf{51.31}$ & $64.52$\\
        \specialrule{0.03em}{0.0pt}{0.8pt}
        \specialrule{0.03em}{0.8pt}{0.0pt}  
        ELECTRA-large & $63.35$ & $90.21$ & $66.59$ & $35.97$ & $83.16$ & $88.78$ & $57.41$ & $70.97$\\
        ELECTRA-our & $\mathbf{66.59}$ & $\mathbf{91.78}$ & $\mathbf{70.12}$ & $\mathbf{39.38}$ & $\mathbf{84.95}$ & $\mathbf{89.83}$ & $\mathbf{58.13}$ & $\mathbf{72.54}$\\
        \specialrule{0.09em}{0.2pt}{0.0pt}
    \end{tabular}
    \caption{Results on classification-style downstream tasks.}
    \label{tab:main_result_1}
\end{table*}

\begin{table*}[tbp]
    \centering
    \small
    \begin{tabular}{l|c|c|c|c|c|c}
        \specialrule{0.09em}{0.0pt}{0.2pt}
        Model & BLEU-1 & BLEU-2 & BLEU-3 & BLEU-4 & METEOR & ROUGE-L\\
        \hline
        BERT & $10.90$ & $3.85$ & $1.69$ & $0.89$ & $4.18$ & $9.80$\\
        GSN & $10.23$ & $3.57$ & $1.70$ & $0.97$ & $4.10$ & $9.91$\\
        HeterMPC$\operatorname{_{BERT}}$ & $\mathbf{12.61}$ & $4.55$ & $2.25$ & $1.41$ & $4.79$ & $11.20$\\
        BERT-our & $11.78$ & $\mathbf{4.74}$ & $\mathbf{2.71}$ & $\mathbf{1.96}$ & $\mathbf{5.09}$ & $\mathbf{11.21}$\\
        \specialrule{0.09em}{0.2pt}{0.0pt}
    \end{tabular}
    \caption{Results on the Ubuntu IRC benchmark.}
    \label{tab:rg_results}
\end{table*}

During the fine-tuning process, we discard the graph-prediction model $q_{\phi}$ since our model no longer requires explicit discourse modeling thanks to the implicit discourse knowledge learn from the pre-training. In our experiments, we make task-specific designs for each downstream task to fully utilize the addressee embedding to lay emphasis on important utterances that are not necessarily addressees, hence we call it \emph{Adaptation Model}. For more details about the task-specific designs, please refer to Appendix \ref{app:adaptation}. To test the universality and simplify the usage of our pre-trained model, experiments are also conducted where we discard the addressee embedding and use only the parameters that are exactly the same as BERT, hence we call it \emph{Vanilla Model}. Following previous works \cite{li-etal-2020-molweni, MPC-BERT, snep}, we mainly conduct our experiments based on BERT$_{\operatorname{base}}$.

In Table \ref{tab:main_result_1}, MPC-BERT \cite{MPC-BERT} is introduced in Section \ref{sec:dialog_pretrain}, which is pre-trained on a small dataset with annotated addressee labels using supervised learning. BERT+CRM is an ablation model that is pre-trained using only the first stage (but with full data), which means only the CRM loss and EM training are adopted. +MLM means addressee-aware MLM objective is further added in the pre-training process and +VI represents our full model with two-stage pre-training. To study whether two-party dialogue models can still work in the multi-party scenario, we also conduct experiments on SPIDER-BERT \cite{zhang-zhao-2021-structural}, which is a model pre-trained on two-party dialogues using self-supervised objectives. 

\subsection{Experimental Results}
We can see from Table \ref{tab:main_result_1} that our full model (+VI) significantly outperforms BERT$_{\operatorname{base}}$ and MPC-BERT on all tasks, justifying the effectiveness of discourse knowledge modeling by incorporating VI into the EM training framework with two-stage pre-training. Besides, BERT+CRM is already strong enough to outperform MPC-BERT or to achieve comparable results, demonstrating the importance of scaling up the pre-training by EM algorithm and incorporating turn-level addressee knowledge. Also, adding addressee-aware MLM adds to performance gains, yet relatively slight. Finally, SPIDER-BERT performs relatively worse than multi-party models, which indicates the significance of designing models and objectives that are specific for multi-party dialogues. For more analyses about why the two-party objectives fail to work on the multi-party scenario, please refer to Appendix \ref{app:two_party}.

Another observation is that the performance drops of the \emph{Vanilla Model} compared with \emph{Adaptation Model} is relatively minor on all dataset, which means it remains powerful even without the task-specific designs. This observation demonstrates that the discourse knowledge is indeed learned and stored in our pre-trained model. 

Besides BERT$_{\operatorname{base}}$, we also experiment with ELECTRA$_{\operatorname{large}}$ to investigate whether our method can still enhance stronger PLMs. In this experiment, we compare the original ELECTRA$_{\operatorname{large}}$ and our full model under the setting of \emph{Adaptation Model}. As shown in the lower part of Table \ref{tab:main_result_1}, our model outperforms ELECTRA$_{\operatorname{large}}$ by large margins. This observation reveals that even strong PLMs, such as ELECTRA$_{\operatorname{large}}$, still lack the knowledge to well understand multi-party dialogues, while our method can effectively enhance them by leveraging the discourse information inferred from the unlabeled datasets.

Our model can also improve the performance of response generation by enhancing the encoder side. Table \ref{tab:rg_results} presents the results on the Ubuntu IRC dataset, where GSN \cite{GSN} and HeterMPC \cite{gu-etal-2022-hetermpc} utilize the discourse annotations in this dataset to explicitly model the reply-to relations by constructing homogeneous or heterogeneous graph neural networks. In contrast, the annotations are not used by our model since it is able to implicitly capture the reply-to information by the discourse knowledge learned during pre-training. As shown in Table \ref{tab:rg_results}, our model outperforms previous models even under the condition that we do not use additional annotations, demonstrating the strong capability of our model to understand the discourse structures.

\section{Analyses}
In this section, we make in-depth analyses to investigate more insights from our method.

\subsection{Ablation Study}
Since our model is trained on massive amounts of data, a natural question is whether the performance gains are from just seeing more conversations. To investigate this, we conduct experiments by removing the addressee-aware EM training process and only performing normal CRM and MLM on the full data. Also to test the out-of-domain generalization ability of our model, for this ablation experiment, we choose SNEP-Twitter and Discourse Parsing tasks since their data sources (Twitter and Ubuntu) are different from our pre-training source (Reddit).

Table \ref{tab:ablation} shows the ablation results, where we observe sharp performance drops when removing the EM training. This observation demonstrates the strong robustness and transferability of our model in out-of-domain data, thanks to the addressee knowledge learned from the EM process.

\subsection{Zero-shot Graph-Prediction}
To investigate to what extent the discourse knowledge is learned by our model, we test the zero-shot graph-prediction task on both Reddit and Molweni datasets. Note that during the pre-training stage, our model is trained on the pseudo-addressee-graphs that are inferred from the unlabeled dataset, hence we call this experiment zero-shot. Table \ref{tab:zero-shot} shows the F1$_\text{G}$ scores of both datasets, where we observe good in-domain performance in Reddit and out-of-domain generalizability in Ubuntu (the Molweni dataset).

\begin{table}[tbp]
    \centering
    \small
    \begin{tabular}{l|cc|cc}
        \specialrule{0.09em}{0.0pt}{0.2pt}
        \multirow{2}{*}{\textbf{Model}} & \multicolumn{2}{c|} {\textbf{Molweni}} & \multicolumn{2}{c} {\textbf{SNEP-Twitter}} \\
        \cline{2-5}
        & F1$_\text{RL}$ & F1$_\text{G}$ & AUC & F1\\
        \hline
        \emph{Adaptation} & & & & \\
        BERT$_\text{+CRM}$ & $63.08$ & $88.40$ & $83.61$ & $89.22$ \\ 
        \quad w/o EM & $61.35$ & $87.69$ & $81.59$ & $88.19$ \\
        BERT$_\text{+CRM+MLM}$ & $63.79$ & $88.42$ & $83.72$ & $89.33$\\
        \quad w/o EM & $61.79$ & $88.04$ & $82.02$ & $88.23$ \\
        \hline
        \emph{Vanilla} & & & & \\
        BERT$_\text{+CRM}$ & $62.95$ & $88.17$ & $82.91$ & $89.11$ \\
        \quad w/o EM & $61.42$ & $88.04$ & $81.45$ & $88.57$ \\
        BERT$_\text{+CRM+MLM}$ & $63.19$ & $88.91$ & $83.48$ & $88.92$\\
        \quad w/o EM & $61.73$ & $88.34$ & $82.12$ & $88.25$ \\
        \specialrule{0.09em}{0.0pt}{0.0pt}
    \end{tabular}
    \caption{Ablation results on the Discourse Parsing (Molweni) and SNEP-Twitter task.}
    \label{tab:ablation}
\end{table}

\begin{table}[tbp]
    \centering
    \small
    \begin{tabular}{l|c|c}
        \specialrule{0.09em}{0.0pt}{0.2pt}
        Model & \textbf{Reddit} & \textbf{Molweni}\\
        \hline
        BERT$_{\operatorname{base}}$ & $74.62$ & $71.94$\\
        ELECTRA$_{\operatorname{large}}$ & $78.71$ & $74.78$\\
        \specialrule{0.09em}{0.2pt}{0.0pt}
    \end{tabular}
    \caption{F1 scores of zero-shot link prediction task.}
    \label{tab:zero-shot}
\end{table}

\begin{figure}[tbp]
    \includegraphics[width=0.46\textwidth]{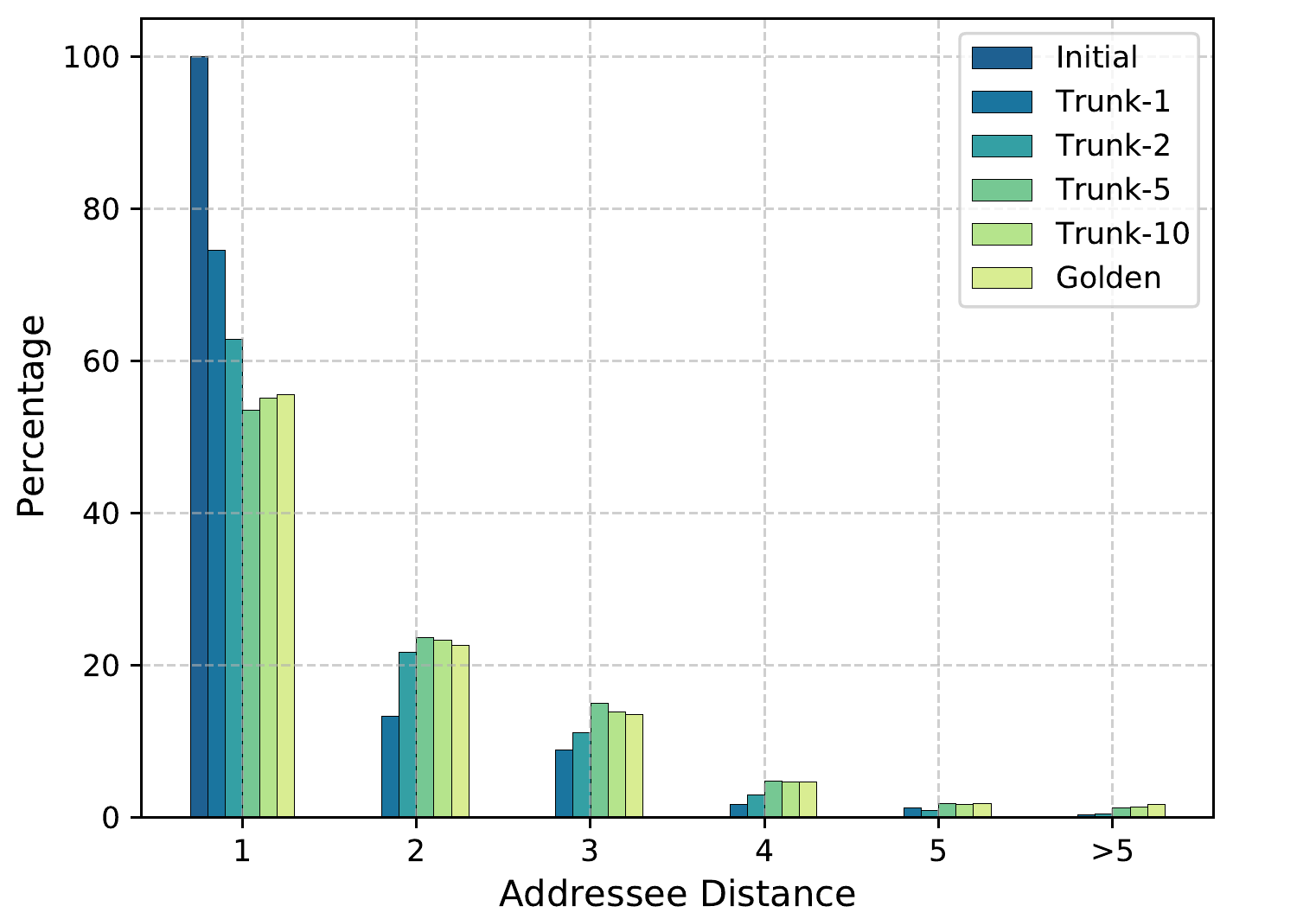}
    \centering
    \caption{Distribution shift of addressee prediction.} 
    \label{fig:dist_shift}
\end{figure}

\begin{figure}[tb]
    \includegraphics[width=0.48\textwidth]{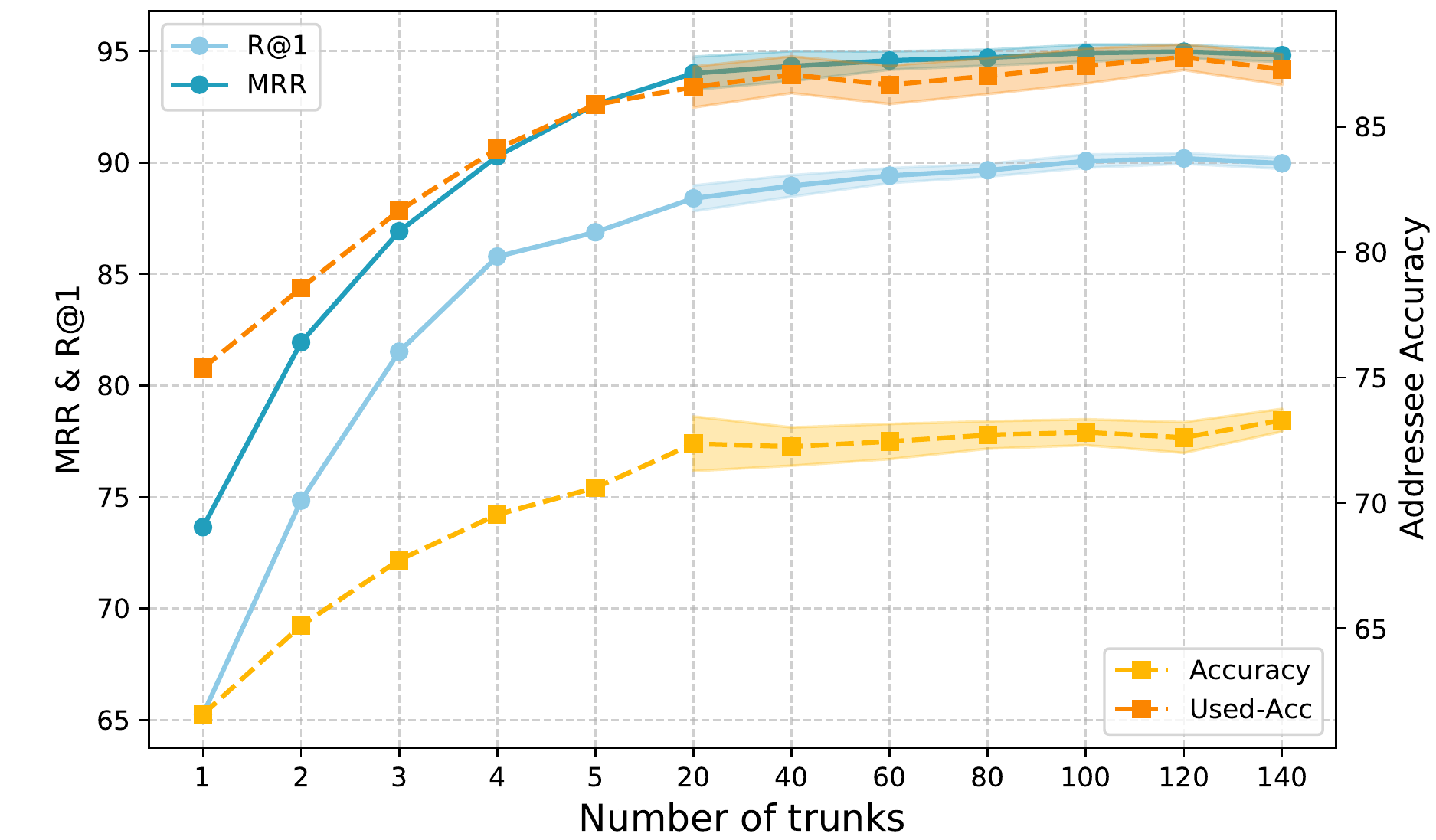}
    \centering
    \caption{CRM scores vs. Addressee prediction accuracy during the pre-training process, where Used-Acc is the accuracy of top $50\%$ confident samples that are used in the next M-step.} 
    \label{fig:em_iterations}
\end{figure}

\subsection{Addressee Distribution Shifts}
\label{sec:dist_shift}
At the beginning of our pre-training process, there are no annotated addressee labels in the training corpus, and the initial model is too weak to infer reasonable addressees using Eq. (\ref{eq:addr_inference}). To cold-start the EM bootstrapping process, we simply set the addressee of every response to be the last utterance in the dialogue history (i.e., $U_{t-1}$), then perform the first round of M-step. This cold-start approach is different from, and much simpler than \citet{li2023em}, where they utilize a trained discourse parser to label the addressees for the first M-step. 

This strategy is simple but exhibits surprisingly good convergence: the distribution of the inferred addressees shifts from one-hot (the initial distribution) to a distribution that is close to the real addressee distribution in an annotated validation set, just after a few trunks. Figure \ref{fig:dist_shift} illustrates the distribution shift, where we draw the validation addressee distance distribution of the last E-step on each trunk. At the initial point, the addressees are all set to the last utterance, hence the percentage of addressees with distance $1$ is $100\%$. With the increase of truck numbers, the addressee distance distribution gradually shifts and becomes closer and closer to the real distribution.

\subsection{Pre-training Trending}
Figure \ref{fig:em_iterations} illustrates the trending of both CRM scores (MRR and Recall@1) and addressee prediction accuracy of ELECTRA$_{\operatorname{large}}$ during the pre-training process. After the $10_{th}$ trunk (the second pre-training stage), we compute the average and standard deviation over the $\pm 10$ trunks of the index and show them in the figure as lines and shades.

First, we can see that both metrics grow together and mutually, which indicates with a stronger CRM model comes better addressee prediction accuracy, demonstrating the correctness of Eq. (\ref{eq:propto}). Besides, the first stage of training reaches its convergence at around the $10_{th}$ trunk, by further incorporating VI at this point, both metrics keep growing and reach their top at around the $120_{th}$ trunk. Finally, the standard deviation is large at the beginning of the second stage of pre-training but gradually decreases with the convergence of the model.

\section{Conclusion}
In this paper, we point out that the lack of annotated addressee labels hinders the scaling-up of multi-party dialogue pre-training. To overcome this obstacle, we propose to utilize the unlabeled datasets by combining the EM algorithm and Variational Inference to jointly infer the discourse labels and pre-train the model with discourse-aware objectives on different granularities. Experimental results and extensive analyses have justified the effectiveness and transferability of our model on multiple downstream tasks.

\section*{Limitations}
Despite the contributions of our work, there are also unavoidable limitations of it.

First, our method is based on the setting that each utterance in the dialogue except the first one has exactly one addressee. This setting holds tightly in online forums such as Twitter or Reddit, yet has its limit in group chats or meetings, where an utterance can reply to multiple or no addressees. However, this scenario is relatively rare in multi-party conversations. Considering this scenario is challenging and complicated since the one-to-many reply-to relations can cause the single-turn EM algorithm intractable. For this part, we leave it to future works.

Second, the Ubuntu IRC benchmark of response generation task is extracted from the Ubuntu Chat Corpus \cite{lowe-etal-2015-ubuntu}, where people discuss the technical issues on the Ubuntu operating system. Due to the lack of human annotators with knowledge of Linux and Ubuntu, we do not conduct human evaluations on this dataset. However, we do provide the generated responses in our supplementary materials for those who are interested in the human evaluations.

\bibliography{anthology,custom}
\bibliographystyle{acl_natbib}

\clearpage

\appendix

\section{Derivation of E-step in Stage-2}
\label{app:estep_stage2}
In the second stage, the maximization step becomes the modeling of $p_{\bm{\theta}}(r_t|Z_t^d, C_{t-1})$, and the expectation step becomes computing the posterior distribution of $p_{\bm{\theta}}(z_t | r_t, Z_{t-1}^d, C_{t-1})$, accordingly. We also factorize this posterior distribution and omit $\bm{\theta}$ for simplicity:
\begin{equation}
    \begin{aligned}
        & p(z_t | r_t, Z_{t-1}^d, C_{t-1}) = \frac{p(z_t, C_{t-1}, Z_{t-1}^d, r_t)}{p(C_{t-1}, r_t, Z_{t-1}^d)}\\
        & = \frac{p(C_{t-1})\ p(Z_{t-1}^d|C_{t-1})\ p(r_t, z_t|C_{t-1}, Z_{t-1}^d)}{p(C_{t-1}) p(Z_{t-1}^d|C_{t-1})\ p(r_t|C_{t-1}, Z_{t-1}^d)}\\
        & = \frac{p(z_t|C_{t-1}, Z_{t-1}^d)\ p(r_t|C_{t-1}, Z_{t-1}^d, z_t)}{p(r_t|C_{t-1},Z_{t-1}^d)}\\
    \end{aligned}
\end{equation}
In this equation, the factorization also follows human habit when we \emph{seriously} participate in a multi-party dialogue: we first read the dialogue history ($C_{t-1}$), then analyze the discourse structure (reply-chains) of it ($Z_{t-1}^d|C_{t-1}$), then choose an addressee utterance we want to reply ($z_t|Z_{t-1}^d, C_{t-1}$), and finally utter a response to it ($r_t|z_t, Z_{t-1}^d, C_{t-1}$). In the last row of this equation, the denominator is irrelevant to $z_t$, and we also assume uniform distribution of $p(z_t|C_{t-1}, Z_{t-1}^d)$ in the numerator, which is also irrelevant to $z_t$. At this point, we can derive that:
\begin{equation}
    p(z_t | r_t, Z_{t-1}^d, C_{t-1}) \propto p(r_t|z_t, Z_{t-1}^d, C_{t-1})
\end{equation}
and calculate the posterior distribution of $z_t$ by traversing all possible addressees $\{z_t^i\}_{i=1}^{t-1}$:
\begin{equation}
    p(z^i_t|r_t, Z_{t-1}^d, C_{t-1}) = \frac{p(r_t|z^i_t, Z_{t-1}^d, C_{t-1})}{\sum \limits_{j=1}^{t-1} p(r_t|z^j_t, Z_{t-1}^d, C_{t-1})}
\end{equation}

\section{Derivation of Prior Distribution}
\label{app:prior}
We now derive how to compute the conditional prior distribution $p_{\bm{\theta}}(Z_t^d|C_{t-1})$, where we also omit $\bm{\theta}$ for simplicity. Firstly, we have
\begin{equation}
    \begin{aligned}
        & p(Z_t^d|C_{t-1}) = p(z_t, Z_{t-1}^d|C_{t-1})\\
        & = p(Z_{t-1}^d|C_{t-1})\ p(z_t|C_{t-1}, Z_{t-1}^d)
    \end{aligned}
\end{equation}
Here $p(z_t|C_{t-1}, Z_{t-1}^d)$ is assumed to be a uniform distribution in Appendix \ref{app:estep_stage2}, so we have:
\begin{equation}
    \label{eq:app_uniform}
    p(z_t|C_{t-1}, Z_{t-1}^d) \sim U(|z_t|)
\end{equation}
where $|z_t|$ is the length of the candidates of $z_t$.
We now focus only on $p(Z_{t-1}^d|C_{t-1})$. Let's note $E(t) = p(z_t | r_t, Z_{t-1}^d, C_{t-1})$, we have:
\begin{equation}
    \label{eq:app_e}
    \begin{aligned}
        & p(Z_{t-1}^d|C_{t-1})\\
        & = p(z_1, z_2,\dots, z_{t-1}|C_{t-1})\\
        & = p(z_1|C_{t-1})\dots p(z_{t-1}|z_1,\dots z_{t-2},C_{t-1})\\
        & = \Pi_{i=1}^{t-1}\ p(z_i|Z_{i-1}^d, C_{t-1})\\
        & = \Pi_{i=1}^{t-1}\ p(z_i|Z_{i-1}^d, C_i)\\
        & = \Pi_{i=1}^{t-1}\ p(z_i|r_i, Z_{i-1}^d, C_{i-1})\\
        & = \Pi_{i=1}^{t-1}\ [E(i)]
    \end{aligned}
\end{equation}
In this equation, we use an intuitive constrain that $p(z_i | Z_{i-1}^d, C_{\ge i}) = p(z_i | Z_{i-1}^d, C_i)$ and $t-1\ge i$, since in real-world scenario, we can not see the future dialogue contexts. Combining Eq. (\ref{eq:app_uniform}) and (\ref{eq:app_e}), we get:
\begin{equation}
    p_{\bm{\theta}}(Z_t^d|C_{t-1}) = \Pi_{i=1}^{t-1} [E(i)]\cdot U(|z_t|)
\end{equation}
which is exactly the same as Eq. (\ref{eq:prior}).

\section{Masked Language Modeling Details}
\label{app:MLM}
For addressee-aware masked language modeling (MLM) object described in Section \ref{sec:total-loss}, the three kinds of special words are masked with a higher probability. Specifically, for normal words, we mask them with a probability of $15\%$, for special words, the probability is $60\%$. The special words are randomly masked first. If the total masking ratio is over $30\%$, we randomly cancel some masks to reduce it below $30\%$. If the total masking ratio is below $15\%$, we repeat the masking process on those normal words to make the final masking ratio from $15\%$ to $30\%$.

\section{Pre-training Details}
\label{app:process}
As mentioned in Section \ref{sec:pretrain-setting}, we split the pre-training data into several trunks and perform EM iterations on each of them. In our experiment, each trunk contains 600,000 ($C_{t-1}$, $r_t^{+/-}$) pairs and the total number of trunks is $158$.

We perform $3$ EM iterations for each trunk. At the end of each trunk, we will load data from the next trunk and perform E-step to infer the initial addressees for the first M-step of the next trunk. Note that the addressee initialization of the first trunk is a heuristic that sets the addressees of all response to the last utterance in the dialogue history, which is mentioned in Section \ref{sec:dist_shift}.

After each E-step, we do not use all the training samples for the next M-step. Instead, we pick the samples with top $50\%$ addressee prediction confidence scores for the next round of M-step. The confidence score is hard to design since simply adopting the highest probability calculated by Eq. (\ref{eq:addr_inference}) will cause length bias: dialogues with shorter context length will have larger highest probability. To solve this issue, we adopt two normalizing methods to normalize the logits output by the model to the same scale, and use the difference between the largest logits and the second largest logits $\operatorname{max} - \operatorname{second\_max}$ to indicate the confidence level. Specifically, the two normalizing methods are min-max normalizing and average normalizing, respectively:
\begin{equation}
    \begin{aligned}
        s_i^{\operatorname{min-max}} &= \frac{s_i - min(S)}{max(S)-min(S)}\\
        s_i^{\operatorname{average}} &= \frac{s_i - min(S)}{avg(S)}
    \end{aligned}
\end{equation}
Here $S = \{s_i\}_{i=1}^{t-1}$ is the logits scores output by the model. For each E-step, we compare the addressee prediction accuracy of the top $50\%$ samples of both normalizing methods in the validation set, then choose the higher one as the normalizing method to select samples for the next round of M-step in the training set. 

To preserve the knowledge learned from the previous trunks and meanwhile fully utilize the newly inferred addressees in each E-step, we remain the parameters of the PLM unchanged and re-initialize the parameters of the addressee embeddings and CRM classifier after each E-step. For the second pre-training stage, we also keep the parameters of the graph-prediction model unchanged.

We start the second stage of pre-training when the vanilla EM algorithm comes to its convergence. Specifically, when the addressee prediction accuracy stops to increase for continuous three trunks, we consider the EM iterations have converged and start the second stage of training by enabling the KL loss and switch the CRM model to the discourse-aware version. In our experiment, the EM algorithm converges at around the $10_{th}$ trunk. In the second stage of pre-training, the hyper-parameters in Eq. (\ref{eq:total-loss}) are set to $\alpha = 1.0$ and $\beta = 0.5$, respectively.

We adopt Simulated Annealing during the Variation Inference to make the pre-training process stable and converge better. Specifically, the temperature coefficient $\tau$ of Eq. (\ref{eq:q-phi}) is set to a high value ($10.0$) at the beginning of the second pre-training stage, then gradually decreases $0.1$ with the graph-prediction model getting stronger and stronger. Formally, in the $i_{th}$ trunk of the second pre-training stage, $\tau$ is calculated as $\tau = \operatorname{max}(0.1, \frac{1}{n-0.9})$.

\section{Dataset Details}
\label{app:datasets}
\textbf{Molweni} is a multi-party dataset for both discourse parsing and question answering tasks. It is sampled from the Ubuntu Chat Corpus \cite{lowe-etal-2015-ubuntu} and is annotated with question-answer pairs and discourse relations (reply-to links and edge types). This dataset contains multi-party dialogues discussing technical issues on the Ubuntu System, hence its topic and domain are very different from our pre-training corpus Reddit. Despite this, our model still generalizes well on this dataset by outperforming the baseline models by large margins. Table \ref{tab:molweni_statistics} shows the statistics of the Molweni dataset, where each utterance is annotated with its addressee and the relation type, each dialogue is annotated with several questions.

\begin{table}[tbp]
    \centering
    \small
    \begin{tabular}{l|c|c|c|c}
        \specialrule{0.09em}{0.0pt}{0.2pt}
        & Train & Dev & Test & Total\\
        \hline
        \# of Dialogues & 8,771 & 883 & 100 & 9,754\\
        \# of Utterances & 77,374 & 7,823 & 845 & 86,042\\
        \# of Questions & 24,682 & 2,513 & 2,871 & 30,066\\
        \specialrule{0.09em}{0.2pt}{0.0pt}
    \end{tabular}
    \caption{Statistic of Molweni dataset.}
    \label{tab:molweni_statistics}
\end{table}

\textbf{Successful New Entry Prediction} (SNEP) is a multi-party dialogue dataset taken from Reddit and Twitter posts. This task is to predict whether a newcomer’s message will be replied to by other users in a multi-party dialogue. This task would be an important part of the research in online assistants and social media. Table \ref{tab:snep_statistics} shows the statistics of the SNEP dataset, where Reddit and Titter are two subsets.

\begin{table}[tbp]
    \centering
    \small
    \begin{tabular}{l|c|c}
        \specialrule{0.09em}{0.0pt}{0.2pt}
        & Twitter & Reddit\\
        \hline
        \# of Dialogues & 37,339 & 69,428\\
        \# of Utterances & 179,265 & 236,764\\
        \# of Questions & 29,340 & 12,199\\
        \# of Successful Entries & 24,682 & 2,513\\
        \# of Failed Entries & 7,999 & 57,229\\
        \specialrule{0.09em}{0.2pt}{0.0pt}
    \end{tabular}
    \caption{Statistic of SNEP dataset.}
    \label{tab:snep_statistics}
\end{table}

\textbf{Ubuntu IRC Benchmark} is a dataset for multi-party dialogue response generation task. This dataset is also from the Ubuntu Chat Corpus \cite{lowe-etal-2015-ubuntu} and contains annotated addressee labels for each utterance. The generation task is formulated as follows: given the dialogue history and a specified addressee, the model should generate an appropriate response that is well related to the addressee. This dataset contains around 380,000 dialogues in total. For developing and testing set, there are 5,000 dialogues, respectively. For the evaluation scripts to compute ROUGE, METEOR, and BLEU, we use the same script as \cite{gu-etal-2022-hetermpc}.

\section{Adaptation Model Details}
\label{app:adaptation}
To make full use of the pre-trained addressee embedding, we design task-specific adaptation method for each downstream task.

For discourse parsing, the use of addressee embedding happens after the reply-to links are predicted. For each reply-to link, we model the addressee (the utterance that is pointed by another) with the addressee embedding and perform the relation classification.

For successful new entry prediction, we infer the addressee of the response to be studied (to predict whether it is a successful new entry) and adopt the addressee embedding to encode the dialogue. We perform mean pooling over the tokens of the response to get a vector, then adopt a binary classifier to make the final prediction.

For extractive question answering, we treat the question ans "response" and the utterance that contains the final answer (key-utterance) span as "addressee". Specifically, during training, we construct key-utterance labels with the annotated answer span and add an auxiliary key-utterance prediction module to predict the key-utterances. We adopt teacher forcing to model the answer span prediction task with the guidance of ground-truth key-utterance information by indicating the key-utterance with the addressee embedding. During inference, we first infer the key-utterance by the key-utterance prediction module, then use the predicted ones to model the answer span prediction task.

\section{Failure of Two-party Objectives}
\label{app:two_party}
Let's take some common objectives of two-party dialogue pre-training for example.  

First, consider the Utterance Order Restoration (UOS) objective that aims to restore the order of permutated utterances in two-party dialogues, or similarly the Utterance Swap Detection (USD) objective that determines whether there exists swapped utterances in the context. In multi-party dialogues, the order of two utterances that reply to the same root-utterance can be swapped, making these two objective inapplicable.

Second, consider the Utterance Restoration and Response Generation/Selection objectives, where the former restores masked utterance tokens using MLM and the latter generates or selects the ground truth response. These objectives can be too difficult for the model to learn without addressee information, due to the one-to-many problem of response-to-context when given different addressees.

The key motivation of this paper and the most difficult part of adopting self-supervised learning on multi-party dialogue is the lack of addressee information, which is subtly addressed by our EM+VI pre-training approach.

\end{document}